\documentclass[letterpaper, 10 pt, conference]{ieeeconf}  

\IEEEoverridecommandlockouts                              

\usepackage{amsmath} 
\usepackage{graphicx}
\usepackage[margin=1in]{geometry} 
\usepackage{tcolorbox}
\usepackage{graphicx}
\usepackage{lipsum}
\usepackage{booktabs}
\usepackage{fancyhdr}
\newcommand{\copyrightheader}{%
    \fancyhf{}
    \fancyhead[C]{\footnotesize
        © 2025 IEEE. Personal use of this material is permitted. Permission from IEEE must be obtained for all other uses, in any current or future media, including reprinting/republishing this material for advertising or promotional purposes, creating new collective works, for resale or redistribution to servers or lists, or reuse of any copyrighted component of this work in other works.\\%
    }%
    \renewcommand{\headrulewidth}{0pt}
}

\title{\LARGE \bf
A Multimodal Pipeline for Clinical Data Extraction: Applying Vision-Language Models to Scans of Transfusion Reaction Reports
}

\author{Henning Schäfer$^{1,2}$,
        Cynthia S. Schmidt$^{1,4}$,
        Johannes Wutzkowsky$^{2}$,
        Kamil Lorek$^{2}$,\\
        Lea Reinartz$^{2}$,
        Johannes Rückert$^{2}$,
        Christian Temme$^{1}$,
        Britta Böckmann$^{2}$,\\
        Peter A. Horn$^{1}$,
        and~Christoph M. Friedrich, \emph{Member,~IEEE}$^{2,3,^*}$%
\thanks{$^{1}$Institute for Transfusion Medicine, University Hospital Essen, 
        Hufelandstraße 55,  Essen, Germany.}%
\thanks{$^{2}$Department of Computer Science, University of Applied Sciences and Arts Dortmund (FHDO), 
        Emil-Figge Str. 42, Dortmund, Germany.}%
\thanks{$^{3}$Institute for Medical Informatics, Biometry and Epidemiology (IMIBE), University Hospital Essen, 
        Hufelandstraße 55, Essen, Germany.}%
\thanks{$^{4}$Institute for AI in Medicine (IKIM), University Hospital Essen, 
        Girardetstraße 2, Essen, Germany.}%
\thanks{$^{5}$Institute of Interventional and Diagnostic Radiology and Neuroradiology, University Hospital Essen, 
        Hufelandstraße 55, Essen, Germany.}%
\thanks{$^{6}$Department of Dermatology, University Hospital Essen,
        Hufelandstraße 55, Essen, Germany.}
\thanks{Contact: \texttt{christoph.friedrich@fh-dortmund.de}}%
}

\begin{document}

\maketitle
\thispagestyle{fancy}   
\copyrightheader        
\pagestyle{empty}

\begin{abstract}
Despite the growing adoption of electronic health records, many processes still rely on paper documents, reflecting the heterogeneous real-world conditions in which healthcare is delivered. The manual transcription process is time-consuming and prone to errors when transferring paper-based data to digital formats. To streamline this workflow, this study presents an open-source pipeline that extracts and categorizes checkbox data from scanned documents. Demonstrated on transfusion reaction reports, the design supports adaptation to other checkbox-rich document types. The proposed method integrates checkbox detection, multilingual optical character recognition (OCR) and multilingual vision-language models (VLMs). The pipeline achieves high precision and recall compared against annually compiled gold-standards from 2017 to 2024. The result is a reduction in administrative workload and accurate regulatory reporting. The open-source availability of this pipeline encourages self-hosted parsing of checkbox forms.
\end{abstract}

\section{INTRODUCTION}
Transfusion reaction reporting is used to assess patient safety and quality assurance within transfusion medicine \cite{ref:transfusion_safety}. Although many healthcare institutions have moved towards fully digital documentation systems \cite{ref:digital_doc}, the degree of digitalization varies widely. Longstanding clinical workflows and preferences often influence the persistence of paper-based documentation \cite{ref:clinical_workflow}. At the German University Hospital taking part in this study, the practical benefits of physically attaching patient- and product-specific stickers onto standardized transfusion reaction forms have led staff to maintain a paper-based approach for transfusion reaction reports. 
This poses challenges when data must be systematically collected and reported \cite{ref:paper_to_digital_challenge}.

One such administrative task is the annual reporting of aggregated transfusion reaction data to national agencies like the Paul Ehrlich Institute (PEI) \cite{ref:pei_requirements}. Historically, the Graduated Plan Officer, charged with compiling these yearly summaries, performed manual data entry into a structured digital format, verifying each entry. This process, although accurate, is laborious, time-consuming, and can introduce errors. It also becomes increasingly difficult to scale as the volume of data continues to grow.

From a clinical perspective, having a searchable, well-structured dataset of transfusion reactions facilitates the identification of trends and patterns, leading to targeted interventions that can improve patient outcomes. 
At the administrative and regulatory level, the collected data can enhance evidence-based research initiatives. 

To support this process, a pipeline is proposed that reads in multilingual checkbox areas on scanned forms into structured data, provided that the categories on the form can be predefined. This pipeline leverages recent advancements in computer vision (CV) and natural language processing (NLP): vision-language models (VLMs) \cite{ref:vlms_general} to understand and interpret the complex visual and textual layout and predefined category mapping as an additional layer to reduce noise and to ensure that clinical findings and suspected diagnoses are accurately classified. While the evaluation focuses on transfusion reaction reports, the techniques are generalizable and can be used by other checkbox-rich paper documents in languages that are supported by Pixtral \cite{journals/corr/abs-2410-07073}.

\begin{figure*}
    \centering
    \includegraphics[width=\linewidth]{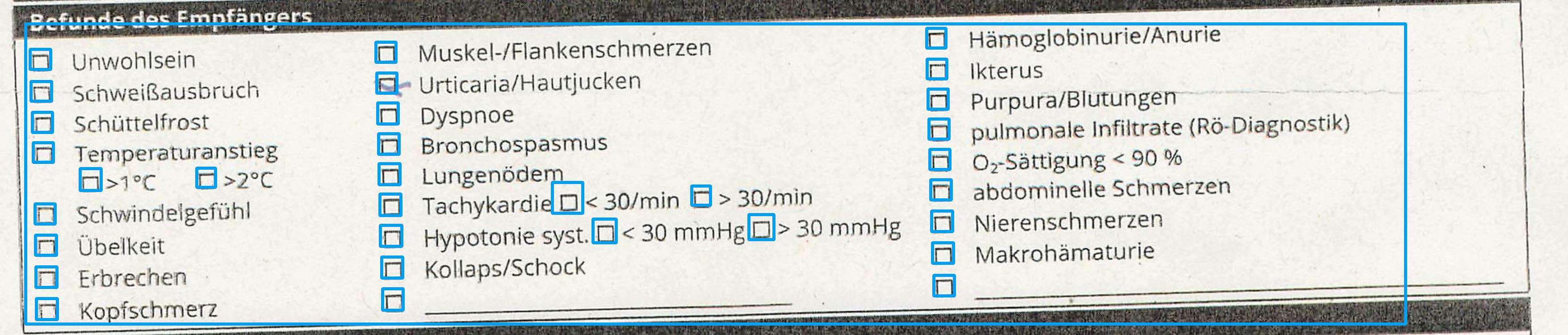}
    \caption{Illustration of the YOLO-based checkbox detection approach applied to a complete transfusion reaction report. Each blue bounding box denotes a detected checkbox area. The model is used to extract the relevant bounding boxes of a contiguous checkbox area to be fed into the Vision Language Model.}
    \label{fig:reactions}
\end{figure*}

Despite the potential of VLMs, automated solutions must contend with the realities of scanning artifacts, varied form layouts, and non-standard placement of checkboxes. Conventional OCR techniques often struggle with these complexities \cite{ref:ocr_limitations}. Automated approaches must demonstrate reliability, transparency, and interpretability to earn the trust of clinical staff who are used to a complete manual verification \cite{ref:trust_requirements}. By building upon recent advances in VLM document parsing, this work addresses the unique challenges of reconstructing transfusion reaction data from scanned documents.

By testing this pipeline on a corpus of transfusion reaction reports spanning 2017 to 2024, it is shown that it can achieve high precision and recall, closely matching the manual workflows. This not only reduces the manual labor and potential errors but also enhances efficiency. 
Additionally, the pipeline is made available as an open-source repository, encouraging broader use, adaptation, and improvement~\footnote{https://github.com/ReMeDi-Blut/Checkbox-Detection-in-Clinical-Documents (Last access: 07.02.25)}.

\subsection{Contributions}
This paper presents a pipeline that addresses these challenges by combining:

\begin{itemize}
    \item \textbf{Checkbox Detection Algorithm}, to identify checkbox areas in documents, enabling subsequent data extraction.
    \item \textbf{Vision-Language Models (VLMs)} , for holistic matching of checked categories within checkbox areas, leveraging both visual and textual understanding to map extracted text onto standardized categories.
\end{itemize}


The remainder of this paper is structured as follows: Section \ref{relwork} introduces related work. Section \ref{data} describes the materials and data used in this study, Section \ref{methods} presents the methods employed, VLM-based checkbox detection, and category mapping, Section \ref{results} reports the results and performance metrics along with qualitative analysis, Section \ref{discussion} discusses the implications, limitations, and potential improvements, and Section \ref{conclusion} concludes with a summary of the contributions and possible future directions.

\section{RELATED WORK}
\label{relwork}
Digitalization of paper-based clinical documents is a longstanding challenge, with OCR being a technology with a long tradition \cite{digitizing_challenge}. OCR systems often relied on handcrafted rules or limited statistical models \cite{journals/pieee/MoriSY92}, struggling with irregular layouts, low image quality, and domain-specific terminologies. Recent work in clinical OCR has shown promise in specialized applications \cite{TIAN2019104840, 9483723, 10222038}. Deep learning-based OCR frameworks, such as Tesseract \cite{TessOverview} with LSTM \cite{journals/neco/HochreiterS97} backends or commercial solutions leveraging convolutional and transformer architectures \cite{conf/nips/VaswaniSPUJGKP17}, have improved accuracy and robustness. Existing solutions still employ template-based approaches or require extensive manual tuning, heavily limiting their generalizability.

\subsection{Vision-Language Models for Document Understanding}
Vision-language models (VLMs) integrate textual and visual signals, enabling a holistic interpretation of images and documents \cite{journals/pami/ZhangHJL24}. Models such as LayoutLM \cite{conf/kdd/XuL0HW020} and LayoutLMv2 \cite{xu-etal-2021-layoutlmv2} use transformer architectures to encode textual content along with spatial layouts, outperforming traditional OCR-only pipelines on form understanding datasets \cite{conf/icdar/JaumeET19}. More recent approaches, e.g., Donut \cite{conf/eccv/KimHYNPYHYHP22} and PaLM-E \cite{conf/icml/DriessXSLCIWTVY23}, utilize image-to-sequence transformers that parse entire document pages directly from pixel inputs without explicit OCR pre-processing, capturing both text and graphical elements in a unified representation.

\subsection{Checkbox Detection and Form Parsing}
Traditional methods rely on image processing heuristics or template matching to identify checked boxes. Modern approaches employ object detection frameworks (e.g., Faster R-CNN \cite{journals/pami/RenHG017}, YOLO \cite{conf/cvpr/RedmonDGF16}, DETR \cite{conf/eccv/CarionMSUKZ20}) trained on annotated datasets to robustly localize checkboxes and classify their states. Beyond checkboxes, form parsing techniques integrate layout analysis and table extraction to reconstruct logical document structures. Recent work has shown that incorporating VLMs can summarize these steps by leveraging semantic context and Visual Question Answering (VQA) Prompting \cite{conf/iccv/AntolALMBZP15}. This speeds up the work compared to creating templates for individual forms.


\section{DATA}
\label{data}

\subsection{Data Source and Characteristics}
The dataset comprises annual transfusion reaction reports collected at a German University Hospital over an eight-year period, spanning from 2017 to 2024. Each entry corresponds to a scanned document that includes various checkboxes for indicating findings and suspected diagnoses as shown in Figure \ref{fig:reactions}. In addition to these categories, the documents contain physical stickers placed on the forms at the time of reporting.

A total of 387 validated transfusion reactions are included. These reactions are associated with 488 transfused blood products, indicating cases where multiple units contributed to the reported reaction. On average, each report contains approximately 1.26 blood products and 3.9 reported findings per reaction. The dataset reflects a balanced gender distribution and covers a broad temporal range, ensuring representative variability in documentation practices. The absolute size remains limited because a record is generated only when staff initiate the protocol for a suspected transfusion reaction, an event with a low incidence rate in routine practice. Over the years, there have been some slight changes to the form, for example the introduction of additional categories, so that larger reference dictionaries had to be configured.

The reports are compiled and aggregated annually by the Graduated Plan Officer, who manually extracts the relevant information to fulfill mandatory reporting obligations. To develop and validate the automated extraction methods, the annual summary tables produced by the Graduated Plan Officer is collected. The data contains all aggregated counts of transfusion reactions and their categorized findings as well as suspected diagnoses and serves as the gold standard that can be compared against extracted data.

Handwritten annotations appeared in approximately 10\% of reaction reports, typically providing additional details for already checked categories (e.g., specifying symptoms for generic "discomfort"). Cases with only handwritten reactions and no corresponding checkbox marks were rare and excluded from the dataset.

\subsection{Ethical and Regulatory Considerations}
All patient-identifying information was handled in accordance with relevant data protection regulations and evaluation was conducted with ethics approval.

\section{METHODS}
\label{methods}
The pipeline processes selected scanned documents from 2017 to 2024, beginning with the validation of blood product and patient identifiers. A YOLOv8-based detection model is then used to identify contiguous checkbox areas within scanned documents (see Figure \ref{fig:reactions}). This model is publicly available and was trained on a custom dataset generated using a Copy-Paste augmentation technique, incorporating document layout analysis to ensure realistic checkbox placement~\footnote{https://github.com/LynnHaDo/Checkbox-Detection (Last access: 07.02.25)}. Training was conducted on over 10,000 synthetic images, with validation performed on 150 human-annotated documents, using YOLO-formatted annotations \cite{harley2015icdar}. Subsequently, two complementary approaches are applied to extract and map the recognized checkboxes to predefined categories, both for findings observed in the recipient (e.g., discomfort, sweating) and suspected diagnoses (e.g., hemolytic reaction). Both approaches were implemented to assess performance in handling common challenges in clinical documentation such as faint or ambiguous check marks, poor scan quality, and varied marking styles.

\subsection{Barcode Detection}
Each transfusion form has a barcode sticker encoding the blood product number. To accurately link reported transfusion reactions to blood products, these barcodes are decoded using pyzbar~\footnote{https://github.com/NaturalHistoryMuseum/pyzbar (Last access: 07.02.25)}. The extracted barcode string includes a 3-digit country code (e.g., “276” for Germany), a 3-digit institute code, and a 9-digit internal serial number. To validate this number, there is a MOD $11.10$ check digit.

\paragraph{Approach 1: OCR-Based Extraction with Levenshtein Matching}
In the first baseline approach, an OCR-driven method is employed. Paddle OCR~\cite{journals/corr/abs-2109-03144} is applied to each text area next to a filled checkbox bounding box detected by the YOLO Model, by a configurable threshold. The OCR is therefore used to extract the text label that is associated with each checked checkbox. To increase the matching probability of OCR outputs to the predefined categories, the Levenshtein distance~\cite{Levenshtein1965BinaryCC} is computed between recognized text and each predefined category. The category with the smallest edit distance is selected as the match for that each filled checkbox.

\paragraph{Approach 2: Vision-Language Model (VLM)-Based Prompting}
Pixtral-Large-Instruct-2411~\cite{journals/corr/abs-2410-07073} is applied to the whole cropped checkbox areas. Instead of processing single checkboxes, the approach focuses specifically on the entire region where checkboxes related to findings or suspected diagnoses are located. This targeted strategy reduces noise and complexity, by enabling the VLM to more effectively interpret and focus on the relevant document excerpt.

To embed more context into the prompt, separate predefined categories are added to instructions used for findings and suspected diagnoses:\\

\begin{tcolorbox}[colback=gray!5!white,colframe=gray!75!black,title=VLM Prompt for Findings of the Recipient]
\textit{You are a medical document analysis assistant to extract the text of marked checkboxes. The following image snippet contains checkboxes from a transfusion reaction report. Each checkbox corresponds to a recipient finding. Possible FINDINGS categories include: [\dots]. Identify which of these categories are checked in the provided image snippet. Do not include any category that is not checked. Provide the checked categories as a simple list.}\\[6pt]
\textbf{User:} [Image snippet of the findings checkboxes attached here]
\end{tcolorbox}

\begin{tcolorbox}[colback=gray!5!white,colframe=gray!75!black,title=VLM Prompt for Suspected Diagnoses]
\textit{You are a medical document analysis assistant to extract the text of marked checkboxes. Each checkbox corresponds to a suspected diagnosis. Possible SUSPECTED DIAGNOSIS categories include: [\dots]. Identify which of these categories are checked in the provided image snippet. Do not include any category that is not checked. Provide the checked categories as a simple list.}\\[6pt]
\textbf{User:} [Image snippet of the suspected diagnoses checkboxes attached here]
\end{tcolorbox}

\section{RESULTS}
\label{results}
A total of 387 transfusion reaction reports from 2017 to 2024 were annually compiled and are available for evaluation purpose of corresponding scans. Each report includes up to 24 possible recipient findings (e.g., fever, discomfort, urticaria) and up to 13 possible suspected diagnoses (e.g., allergic reaction, hemolytic reaction, TRALI).

\subsection{Barcode Detection}
Table~\ref{tab:barcode_detection_validation} summarizes the performance of barcode extraction.

\begin{table}[htbp]
    \centering
    \caption{Barcode Detection and Validation Results}
    \begin{tabular}{l c}
        \toprule
        \textbf{Metric} & \textbf{Value} \\
        \midrule
        Blood Product Barcode Accuracy & 93.21\%  \\
        Patient Sticker Barcode Accuracy & 89.75\%  \\
        \bottomrule
    \end{tabular}
    \label{tab:barcode_detection_validation}
\end{table}

While these results on barcode detection do not impact the checkbox detection, they are still reported here to provide a real-world assessment of barcode detection accuracy for blood product and patient stickers on scans.

\subsection{Findings and Suspected Diagnosis Category Mapping}
Two approaches for extracting and mapping these checked categories to predefined labels were compared:

\begin{itemize}
    \item \textbf{OCR + Levenshtein Matching:}
        PaddleOCR is used to read text directing horizontally from each filled checkbox region detected by the YOLO model, and the extracted text is then matched to the closest predefined category using Levenshtein distance.
    
    \item \textbf{VLM-Based Prompting:}
        A vision-language model (Pixtral-Large-Instruct-2411) is given the image snippet of the checkbox area together with a list of known categories; it infers which categories are checked based on the prompts.
\end{itemize}

Table~\ref{tab:ocr_result_summary} summarizes the OCR-based approach. While edit-distance matching helps correct minor text distortions, some scan samples raise the risk of confusion finding the correct category if the OCR output is heavily corrupted or truncated.

\begin{table}[htbp]
    \centering
    \caption{OCR + Levenshtein Category Extraction}
    \begin{tabular}{l c c c}
        \toprule
        \textbf{Category Set} & \textbf{Precision} & \textbf{Recall} & \textbf{F1-Score}\\
        \midrule
        Findings (24) & 88.27\% & 84.39\% & 86.27\% \\
        Suspected Diagnoses (13) & 89.04\% & 85.46\% & 87.21\% \\
        \bottomrule
    \end{tabular}
    \label{tab:ocr_result_summary}
\end{table}

Table~\ref{tab:vlm_result_summary} presents the performance of the VLM-based method. Despite the number of categories, it maintains strong robustness against faint or partially marked checkboxes and mild scan distortions.

\begin{table}[htbp]
    \centering
    \caption{VLM-Based Category Extraction}
    \begin{tabular}{l c c c}
        \toprule
        \textbf{Category Set} & \textbf{Precision} & \textbf{Recall} & \textbf{F1-Score}\\
        \midrule
        Findings (24) & 93.21\% & 89.24\% & 91.18\% \\
        Suspected Diagnoses (13) & 94.08\% & 91.64\% & 92.84\% \\
        \bottomrule
    \end{tabular}
    \label{tab:vlm_result_summary}
\end{table}

To provide an aggregated view across both findings and suspected diagnoses, Table~\ref{tab:category_mapping_accuracy} reports the average accuracy for each approach.

\begin{table}[htbp]
    \centering
    \caption{Category Mapping Accuracy: VLM vs. OCR + Levenshtein}
    \begin{tabular}{l c c}
        \toprule
        \textbf{Approach} & \textbf{Accuracy (Avg)} \\
        \midrule
        VLM-Based          & 92.04\%  \\
        OCR + Levenshtein  & 85.17\%  \\
        \bottomrule
    \end{tabular}
    \label{tab:category_mapping_accuracy}
\end{table}

\section{DISCUSSION}
\label{discussion}
The results indicate that the proposed VLM-based approach consistently outperforms the OCR + Levenshtein baseline for scanned transfusion reaction forms. The high accuracy of the vision-language pipeline for category mapping suggests promising directions for broader adoption. Three primary discussion points that surfaced during evaluation:

\begin{itemize}
    \item \textbf{Partial Check Marks:} 
    The VLM-based method demonstrate its capacity to interpret faint or incomplete check marks. By leveraging holistic image analysis rather than relying solely on a fill threshold for the detected checkbox region, the model can identify visual hints of a checked box even under suboptimal conditions. This capability is of particular importance, where rapid form-filling and different practitioners can introduce inconsistencies in checking behaviors. An analysis of 24 forms with faint, corrected or ambiguous markings showed VLM correctly identified 21 (87.5\%) compared to OCR's 16 (66.7\%).

    \item \textbf{Expanded Category Set:} 
    With up to 24 recipient findings and 13 suspected diagnoses, the volume of potential categories increased the difficulty for both methods. The analysis showed that the VLM approach remains robust despite long context prompts that can introduce a risk of category confusion when a large number of options must be distinguished.

    \item \textbf{Poor Scan Quality and Artifacts:}
    Another challenge arose in scans that were degraded e.g., smudges and low resolution. These conditions led to considerable failures in OCR recognition. In many of these difficult cases, the VLM-based approach was able to infer the correct selection by exploiting whatever visual and textual information on the scan remained intact. This suggests that VLMs can add a layer of resilience to document parsing workflows, especially in real-world clinical settings where scan quality cannot be guaranteed.
    
\end{itemize}

In addition to its use in digitalization, the application can also be directly integrated into transfusion medicine workflows that require a mandatory dual control principle to ensure verification by at least two human operators. The system can act as a third level of validation here by detecting discrepancies and initiating a review in ambiguous cases to further enhance safety. In scenarios where infrastructure is unavailable or limited, such as mass casualty incidents or system failures, the ability to quickly extract data from paper forms supports operational continuity.

\subsection{Limitations}
\label{limitations}
While the performance metrics are strong, some limitations should be acknowledged. The accuracy still depends on scan quality and the clarity of checkbox markings. Documents with heavily distorted text, or unusual labeling practices may always require manual review. Although such cases are rare, they remain a source of potential error. No instances of misclassifying severe reactions (e.g., anaphylaxis) as mild were observed, though such errors would have great clinical implications and require intervention by human-in-the-loop verification. A further challenge arises from the presence of optional handwritten entries, which was not part of this evaluation study: Although both approaches can detect handwritten text to some extent, multilingual jargon-heavy and written notes by practitioners working under constant time pressure is extremely hard to detect. 

Though the VLM approach may work on a variety of documents, the current configuration is evaluated by one type of form. Extending this solution to other institutions may require adjusting the detection and classification components to adjust for different layouts and category nomenclatures. While the methods are generalizable, applying them elsewhere might involve fine-tuning the models, adding custom dictionaries, and modifying a few parts of the source code. 

\section{Conclusion}
\label{conclusion}
This work presents a technically robust, fully automated pipeline for extracting and categorizing transfusion reaction data from paper-based scans. The open source pipeline convert checkbox rich documents into structured, machine-readable outputs. Within the conducted evaluation, the resulting data aligns with the annually compiled gold-standard datasets created by the Graduated Plan Officer.

Here, the VLM-based pipeline demonstrates strong potential in reducing manual data entry burdens and improving reporting accuracy for transfusion reactions. However, scaling to more complex forms and ensuring reliable performance under image distortions remain areas for further investigation. Future work will focus on integrating more advanced error-correction techniques, aiming toward a fully automated, end-to-end solution to import the structured data into the hospital information system, so that transfusion reactions can be linked to the rest of the patient data, which allows for tasks such as predicting the transfusion outcome with regard to possible risks of adverse reactions and intolerances.

The open-source approach encourages customization and extension over proprietary integrations, such as adapting to other clinical forms and medical reporting tasks. The proposed self-hosted workflow serves as a blueprint for open and adaptable document processing solutions in healthcare validated by a real-world scenario, hopefully leading to improved quality management and data-driven clinical data collection.



\bibliographystyle{IEEEtran}
\bibliography{bibliography}

\begin{thebibliography}{10}
\providecommand{\url}[1]{#1}
\csname url@samestyle\endcsname
\providecommand{\newblock}{\relax}
\providecommand{\bibinfo}[2]{#2}
\providecommand{\BIBentrySTDinterwordspacing}{\spaceskip=0pt\relax}
\providecommand{\BIBentryALTinterwordstretchfactor}{4}
\providecommand{\BIBentryALTinterwordspacing}{\spaceskip=\fontdimen2\font plus
\BIBentryALTinterwordstretchfactor\fontdimen3\font minus \fontdimen4\font\relax}
\providecommand{\BIBforeignlanguage}[2]{{%
\expandafter\ifx\csname l@#1\endcsname\relax
\typeout{** WARNING: IEEEtran.bst: No hyphenation pattern has been}%
\typeout{** loaded for the language `#1'. Using the pattern for}%
\typeout{** the default language instead.}%
\else
\language=\csname l@#1\endcsname
\fi
#2}}
\providecommand{\BIBdecl}{\relax}
\BIBdecl

\bibitem{ref:transfusion_safety}
R.~R.~P. de~Vries, J.-C. Faber, P.~F.~W. Strengers, and M.~of~the Board of~the International Haemovigilance~Network, ``{Haemovigilance: an effective tool for improving transfusion practice},'' \emph{Vox Sanguinis}, vol. 100, no.~1, pp. 60--67, 2011.

\bibitem{ref:digital_doc}
L.~A. Baumann, J.~Baker, and A.~G. Elshaug, ``{The impact of electronic health record systems on clinical documentation times: A systematic review},'' \emph{Health Policy}, vol. 122, no.~8, pp. 827--836, 2018.

\bibitem{ref:clinical_workflow}
J.~J. Saleem, A.~L. Russ, C.~F. Justice, H.~Hagg, P.~R. Ebright, P.~A. Woodbridge, and B.~N. Doebbeling, ``{Exploring the persistence of paper with the electronic health record},'' \emph{International Journal of Medical Informatics}, vol.~78, no.~9, pp. 618--628, 2009.

\bibitem{ref:paper_to_digital_challenge}
U.~Sivarajah, M.~M. Kamal, Z.~Irani, and V.~Weerakkody, ``{Critical analysis of Big Data challenges and analytical methods},'' \emph{Journal of Business Research}, vol.~70, pp. 263--286, 2017.

\bibitem{ref:pei_requirements}
B.~Keller-Stanislawski, A.~Lohmann, S.~Günay, M.~Heiden, and M.~B. Funk, ``{The German Haemovigilance System–reports of serious adverse transfusion reactions between 1997 and 2007},'' \emph{Transfusion Medicine}, vol.~19, no.~6, pp. 340--349, 2009.

\bibitem{ref:vlms_general}
J.~Zhang, J.~Huang, S.~Jin, and S.~Lu, ``{Vision-Language Models for Vision Tasks: A Survey},'' \emph{IEEE Transactions on Pattern Analysis and Machine Intelligence}, vol.~46, no.~8, pp. 5625--5644, 2024.

\bibitem{journals/corr/abs-2410-07073}
P.~Agrawal, S.~Antoniak, E.~B. Hanna, B.~Bout, D.~S. Chaplot, J.~Chudnovsky, D.~Costa, B.~D. Monicault, S.~Garg, T.~Gervet, S.~Ghosh, A.~H{\'{e}}liou, P.~Jacob, A.~Q. Jiang, K.~Khandelwal, T.~Lacroix, G.~Lample, D.~de~Las~Casas, T.~Lavril, T.~L. Scao, A.~Lo, W.~Marshall, L.~Martin, A.~Mensch, P.~Muddireddy, V.~Nemychnikova, M.~Pellat, P.~von Platen, N.~Raghuraman, B.~Rozi{\`{e}}re, A.~Sablayrolles, L.~Saulnier, R.~Sauvestre, W.~Shang, R.~Soletskyi, L.~Stewart, P.~Stock, J.~Studnia, S.~Subramanian, S.~Vaze, T.~Wang, and S.~Yang, ``{Pixtral 12B},'' \emph{CoRR}, vol. abs/2410.07073 v2, 2024.

\bibitem{ref:ocr_limitations}
T.~Sato, T.~Kanade, E.~K. Hughes, M.~A. Smith, and S.~Satoh, ``{"Video OCR: indexing digital news libraries by recognition of superimposed captions"},'' \emph{Multimedia Systems}, vol.~7, no.~5, pp. 385--395, Sep 1999.

\bibitem{ref:trust_requirements}
A.~Chavaillaz, D.~Wastell, and J.~Sauer, ``{System reliability, performance and trust in adaptable automation},'' \emph{Applied Ergonomics}, vol.~52, pp. 333--342, 2016.

\bibitem{digitizing_challenge}
C.~Thorat, A.~Bhat, P.~Sawant, I.~Bartakke, and S.~Shirsath, ``{"A Detailed Review on Text Extraction Using Optical Character Recognition"},'' in \emph{ICT Analysis and Applications}, S.~Fong, N.~Dey, and A.~Joshi, Eds.\hskip 1em plus 0.5em minus 0.4em\relax Singapore: Springer Nature Singapore, 2022, pp. 719--728.

\bibitem{journals/pieee/MoriSY92}
S.~Mori, C.~Y. Suen, and K.~Yamamoto, ``{Historical review of {OCR} research and development},'' \emph{Proc. {IEEE}}, vol.~80, no.~7, pp. 1029--1058, 1992.

\bibitem{TIAN2019104840}
Q.~Tian, M.~Liu, L.~Min, J.~An, X.~Lu, and H.~Duan, ``{An automated data verification approach for improving data quality in a clinical registry},'' \emph{Computer Methods and Programs in Biomedicine}, vol. 181, p. 104840, 2019, sI: Data Quality Assessment.

\bibitem{9483723}
E.~Murphy, S.~Samuel, J.~Cho, W.~Adorno, M.~Durieux, D.~Brown, and C.~Ndaribitse, ``{Checkbox Detection on Rwandan Perioperative Flowsheets using Convolutional Neural Network},'' in \emph{2021 Systems and Information Engineering Design Symposium (SIEDS)}, 2021, pp. 1--6.

\bibitem{10222038}
M.~A. Zaryab and C.~R. Ng, ``{Optical Character Recognition for Medical Records Digitization with Deep Learning},'' in \emph{2023 IEEE International Conference on Image Processing (ICIP)}, 2023, pp. 3260--3263.

\bibitem{TessOverview}
R.~Smith, ``{An Overview of the Tesseract OCR Engine},'' in \emph{ICDAR '07: Proceedings of the Ninth International Conference on Document Analysis and Recognition}.\hskip 1em plus 0.5em minus 0.4em\relax Washington, DC, USA: IEEE Computer Society, 2007, pp. 629--633.

\bibitem{journals/neco/HochreiterS97}
S.~Hochreiter and J.~Schmidhuber, ``{Long Short-Term Memory},'' \emph{Neural Comput.}, vol.~9, no.~8, pp. 1735--1780, 1997.

\bibitem{conf/nips/VaswaniSPUJGKP17}
A.~Vaswani, N.~Shazeer, N.~Parmar, J.~Uszkoreit, L.~Jones, A.~N. Gomez, L.~Kaiser, and I.~Polosukhin, ``{Attention is All you Need},'' in \emph{Advances in Neural Information Processing Systems 30: Annual Conference on Neural Information Processing Systems 2017, December 4-9, 2017, Long Beach, CA, {USA}}, I.~Guyon, U.~von Luxburg, S.~Bengio, H.~M. Wallach, R.~Fergus, S.~V.~N. Vishwanathan, and R.~Garnett, Eds., 2017, pp. 5998--6008.

\bibitem{journals/pami/ZhangHJL24}
J.~Zhang, J.~Huang, S.~Jin, and S.~Lu, ``{Vision-Language Models for Vision Tasks: {A} Survey},'' \emph{{IEEE} Trans. Pattern Anal. Mach. Intell.}, vol.~46, no.~8, pp. 5625--5644, 2024.

\bibitem{conf/kdd/XuL0HW020}
Y.~Xu, M.~Li, L.~Cui, S.~Huang, F.~Wei, and M.~Zhou, ``{LayoutLM: Pre-training of Text and Layout for Document Image Understanding},'' in \emph{{KDD} '20: The 26th {ACM} {SIGKDD} Conference on Knowledge Discovery and Data Mining, Virtual Event, CA, USA, August 23-27, 2020}, R.~Gupta, Y.~Liu, J.~Tang, and B.~A. Prakash, Eds.\hskip 1em plus 0.5em minus 0.4em\relax {ACM}, 2020, pp. 1192--1200.

\bibitem{xu-etal-2021-layoutlmv2}
Y.~Xu, Y.~Xu, T.~Lv, L.~Cui, F.~Wei, G.~Wang, Y.~Lu, D.~Florencio, C.~Zhang, W.~Che, M.~Zhang, and L.~Zhou, ``{"{L}ayout{LM}v2: Multi-modal Pre-training for Visually-rich Document Understanding"},'' in \emph{Proceedings of the 59th Annual Meeting of the Association for Computational Linguistics and the 11th International Joint Conference on Natural Language Processing (Volume 1: Long Papers)}, C.~Zong, F.~Xia, W.~Li, and R.~Navigli, Eds.\hskip 1em plus 0.5em minus 0.4em\relax Online: Association for Computational Linguistics, Aug. 2021, pp. 2579--2591.

\bibitem{conf/icdar/JaumeET19}
G.~Jaume, H.~K. Ekenel, and J.~Thiran, ``{{FUNSD:} {A} Dataset for Form Understanding in Noisy Scanned Documents},'' in \emph{2nd International Workshop on Open Services and Tools for Document Analysis, OST@ICDAR 2019, Sydney, Australia, September 22-25, 2019}.\hskip 1em plus 0.5em minus 0.4em\relax {IEEE}, 2019, pp. 1--6.

\bibitem{conf/eccv/KimHYNPYHYHP22}
G.~Kim, T.~Hong, M.~Yim, J.~Nam, J.~Park, J.~Yim, W.~Hwang, S.~Yun, D.~Han, and S.~Park, ``{OCR-Free Document Understanding Transformer},'' in \emph{Computer Vision - {ECCV} 2022 - 17th European Conference, Tel Aviv, Israel, October 23-27, 2022, Proceedings, Part {XXVIII}}, ser. Lecture Notes in Computer Science, S.~Avidan, G.~J. Brostow, M.~Ciss{\'{e}}, G.~M. Farinella, and T.~Hassner, Eds., vol. 13688.\hskip 1em plus 0.5em minus 0.4em\relax Springer, 2022, pp. 498--517.

\bibitem{conf/icml/DriessXSLCIWTVY23}
D.~Driess, F.~Xia, M.~S.~M. Sajjadi, C.~Lynch, A.~Chowdhery, B.~Ichter, A.~Wahid, J.~Tompson, Q.~Vuong, T.~Yu, W.~Huang, Y.~Chebotar, P.~Sermanet, D.~Duckworth, S.~Levine, V.~Vanhoucke, K.~Hausman, M.~Toussaint, K.~Greff, A.~Zeng, I.~Mordatch, and P.~Florence, ``{PaLM-E: An Embodied Multimodal Language Model},'' in \emph{International Conference on Machine Learning, {ICML} 2023, 23-29 July 2023, Honolulu, Hawaii, {USA}}, ser. Proceedings of Machine Learning Research, A.~Krause, E.~Brunskill, K.~Cho, B.~Engelhardt, S.~Sabato, and J.~Scarlett, Eds., vol. 202.\hskip 1em plus 0.5em minus 0.4em\relax {PMLR}, 2023, pp. 8469--8488.

\bibitem{journals/pami/RenHG017}
S.~Ren, K.~He, R.~B. Girshick, and J.~Sun, ``{Faster {R-CNN:} Towards Real-Time Object Detection with Region Proposal Networks},'' \emph{{IEEE} Trans. Pattern Anal. Mach. Intell.}, vol.~39, no.~6, pp. 1137--1149, 2017.

\bibitem{conf/cvpr/RedmonDGF16}
J.~Redmon, S.~K. Divvala, R.~B. Girshick, and A.~Farhadi, ``{You Only Look Once: Unified, Real-Time Object Detection},'' in \emph{2016 {IEEE} Conference on Computer Vision and Pattern Recognition, {CVPR} 2016, Las Vegas, NV, USA, June 27-30, 2016}.\hskip 1em plus 0.5em minus 0.4em\relax {IEEE} Computer Society, 2016, pp. 779--788.

\bibitem{conf/eccv/CarionMSUKZ20}
N.~Carion, F.~Massa, G.~Synnaeve, N.~Usunier, A.~Kirillov, and S.~Zagoruyko, ``{End-to-End Object Detection with Transformers},'' in \emph{Computer Vision - {ECCV} 2020 - 16th European Conference, Glasgow, UK, August 23-28, 2020, Proceedings, Part {I}}, ser. Lecture Notes in Computer Science, A.~Vedaldi, H.~Bischof, T.~Brox, and J.~Frahm, Eds., vol. 12346.\hskip 1em plus 0.5em minus 0.4em\relax Springer, 2020, pp. 213--229.

\bibitem{conf/iccv/AntolALMBZP15}
S.~Antol, A.~Agrawal, J.~Lu, M.~Mitchell, D.~Batra, C.~L. Zitnick, and D.~Parikh, ``{VQA:} visual question answering,'' in \emph{2015 {IEEE} International Conference on Computer Vision, {ICCV} 2015, Santiago, Chile, December 7-13, 2015}.\hskip 1em plus 0.5em minus 0.4em\relax {IEEE} Computer Society, 2015, pp. 2425--2433.

\bibitem{harley2015icdar}
A.~W. Harley, A.~Ufkes, and K.~G. Derpanis, ``{Evaluation of Deep Convolutional Nets for Document Image Classification and Retrieval},'' in \emph{{International Conference on Document Analysis and Recognition ({ICDAR})}}, 2015.

\bibitem{journals/corr/abs-2109-03144}
Y.~Du, C.~Li, R.~Guo, C.~Cui, W.~Liu, J.~Zhou, B.~Lu, Y.~Yang, Q.~Liu, X.~Hu, D.~Yu, and Y.~Ma, ``{PP-OCRv2: Bag of Tricks for Ultra Lightweight {OCR} System},'' \emph{CoRR}, vol. abs/2109.03144 v2, 2021.

\bibitem{Levenshtein1965BinaryCC}
V.~I. Levenshtein, ``Binary codes capable of correcting deletions, insertions, and reversals,'' \emph{Soviet physics. Doklady}, vol.~10, pp. 707--710, 1965.

\end{thebibliography}

\end{document}